\documentclass{ws-procs11x85}
\usepackage{ws-procs-thm} 
\usepackage{enumitem}
\usepackage{hyperref}

\usepackage{algorithm}
\usepackage{algpseudocode}

\usepackage[most]{tcolorbox}

\newcommand\ci{\perp\!\!\!\perp}

\DeclareMathOperator*{\argmax}{arg\,max}

\begin{document}
\title{Learning Causally Predictable Outcomes from Psychiatric Longitudinal Data}

\author{Eric V. Strobl}

\address{Departments of Biomedical Informatics \& Psychiatry, University of Pittsburgh,\\
Pittsburgh, Pennsylvania 15206, United States of America}

\begin{abstract}
Causal inference in longitudinal biomedical data remains a central challenge, especially in psychiatry, where symptom heterogeneity and latent confounding frequently undermine classical estimators. Most existing methods for treatment effect estimation presuppose a fixed outcome variable and address confounding through observed covariate adjustment. However, the assumption of unconfoundedness may not hold for a fixed outcome in practice. To address this foundational limitation, we directly optimize the outcome definition to maximize causal identifiability. Our DEBIAS (Durable Effects with Backdoor-Invariant Aggregated Symptoms) algorithm learns non-negative, clinically interpretable weights for outcome aggregation, maximizing durable treatment effects and empirically minimizing both observed and latent confounding by leveraging the time-limited direct effects of prior treatments in psychiatric longitudinal data. The algorithm also furnishes an empirically verifiable test for outcome unconfoundedness. DEBIAS consistently outperforms state-of-the-art methods in recovering causal effects for clinically interpretable composite outcomes across comprehensive experiments in depression and schizophrenia. R code is available at \url{github.com/ericstrobl/DEBIAS}.
\end{abstract}


% required, do-not-remove
%\copyrightinfo{\copyright\ 2025 The Authors. Open Access chapter published by World Scientific Publishing Company and distributed under the terms of the Creative Commons Attribution Non-Commercial (CC BY-NC) 4.0 License.}

\section{Introduction}\label{sec:intro}

Causal inference seeks to identify cause-and-effect relationships from data, enabling scientific discovery and effective intervention design \cite{Imbens15,Spirtes00}. In psychiatry, most causal knowledge has come from randomized clinical trials (RCTs), but RCTs are often infeasible or unethical to perform. For example, we cannot randomize individuals to different income levels to study the effects of poverty on mental health. Meanwhile, rich longitudinal observational datasets that provide far greater temporal and phenotypic resolution than cross-sectional studies \cite{Hedeker06} or electronic health records \cite{Goldstein16} are increasingly accessible through protected repositories in psychiatry. However, inferring causality from such data remains difficult in the absence of randomized assignment. 

\begin{figure}[t]
\centering
\includegraphics[width=0.75\textwidth]{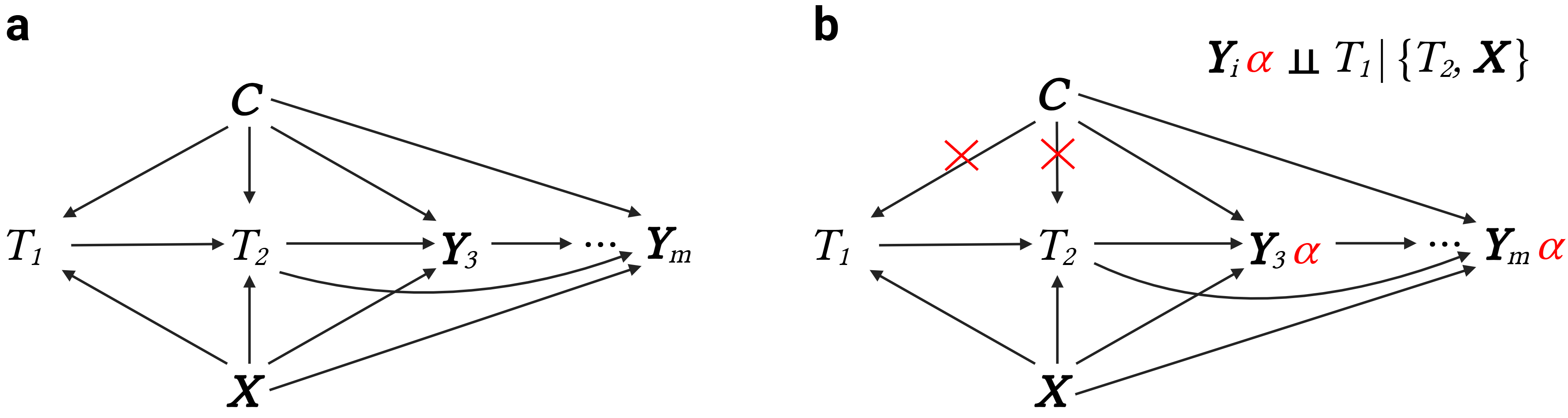}
\caption{ \textbf{Main idea.} (a) Causal diagram where $T_2$ denotes current treatment assignment, $T_1$ denotes historical treatment assignment (not necessarily representing the same treatments as in $T_2$), each $\bm{Y}_i$ represents multiple items of a clinical rating scale, $\bm{X}$ indicates observed confounders available for adjustment, and $\bm{C}$ denotes latent confounders. In this structure, $T_1$ has time-limited direct effects: its influence on $\bm{Y}_3, \ldots, \bm{Y}_m$ is entirely indirect -- mediated through $T_2$ -- with no direct arrows from $T_1$ to the later outcomes. Moreover, $\bm{Y}_i$ is either unweighted or combined in a non-optimized manner, such as using total severity scores. 
(b) We learn non-negative weights $\alpha \geq 0$ to form the weighted severity scores $\bm{Y}_3\alpha, \dots, \bm{Y}_m \alpha$ such that $\bm{Y}_i\alpha \ci T_1 \mid \{T_2, \bm{X}\}$ for each time step. This outcome projection removes the statistical influence of spurious backdoor associations between $T_1$ and the outcomes, as indicated by the red crosses.}
\label{fig:main}
\end{figure}

One of the primary obstacles to causal inference from observational data is latent confounding \cite{Imbens15}, where unmeasured variables $\bm{C}$ may influence both treatment $T_2$ and outcomes ($\bm{Y}_3, \ldots, \bm{Y}_m$), biasing causal effect estimates (Figure~\ref{fig:main} (a)). Here, subscripts denote time steps in the longitudinal data, so $T_1$ and $T_2$ refer to treatments at different times, and $\bm{Y}_i$ represents the vector of \textit{multiple} clinical rating scale items measured at time $i$. While standard practice adjusts for observed confounders $\bm{X}$ in regression models, this ignores clinical knowledge that many historical treatments $T_1$ have only \textit{time-limited direct causal effects} and thus do not directly impact later outcomes\footnote{We relax this assumption in Supplementary Materials \ref{SM:relax}, where we also provide more technical discussion.}. Instead, any effect of $T_1$ on outcomes $\bm{Y}_3, \ldots, \bm{Y}_m$ occurs indirectly -- mediated through $T_2$ -- with no direct causal arrows from $T_1$ to the later outcomes (Figure \ref{fig:main} (a)). For instance, antidepressants do not directly affect symptom severity a year after discontinuation in active major depression \cite{Rush06a}.

We propose to incorporate the prior knowledge of short-term treatment effects by searching for a non-negative weight vector $\alpha$ such that the aggregated outcome $\bm{Y}_i\alpha$ is conditionally independent of past treatment $T_1$, given current treatment $T_2$ and $\bm{X}$, for all time steps. The non-negativity constraint on $\alpha$ ensures that $\bm{Y}_i\alpha$ remains a clinically interpretable \textit{severity score}, preserving the additive contribution of each symptom without introducing cancellation effects. Moreover, in this longitudinal setting, the latent confounders linking $T_1$ to $\bm{Y}_i\alpha$ are assumed to be the same as those linking $T_2$ to $\bm{Y}_i\alpha$, since both $T_1$ and $T_2$ represent treatment assignments differing by time. As a result, achieving conditional independence between $T_1$ and $\bm{Y}_i\alpha$ (after adjusting for $T_2$ and $\bm{X}$) eliminates confounding for both past and current treatments. Graphically, this corresponds to eliminating the statistical influence transmitted along non-causal paths between $T_1$ and each $\bm{Y}_i\alpha$, thereby enabling unbiased inference of the causal effect of $T_2$ on each $\bm{Y}_i\alpha$ (e.g., Figure~\ref{fig:main} (b)). Applying the same $\alpha$ across all time points ultimately yields a single, durable, and empirically unconfounded severity score, enabling credible causal inference for a structured combination of items -- even when such inference is infeasible for every individual item.

\begin{tcolorbox}[enhanced,frame hidden,breakable]
We specifically make the following contributions in this work:
\begin{enumerate}[leftmargin=*]
\item We introduce a principled approach for learning outcome scores as non-negative combinations of clinical rating scale items, maximizing the correlation between current treatment and subsequent outcomes across time.
\item We propose a novel regularization criterion that minimizes latent confounding bias by leveraging the time-limited direct effects of past treatments.
\item We generalize the framework to extract not just a single score, but all unconfounded severity scores, each defined by a distinct non-negative weight vector and interpretable as a composite severity score.
\item We instantiate the above concepts into a single algorithm, the Durable Effects with Backdoor-Invariant Aggregated Symptoms (DEBIAS), which employs cross-validation to minimize latent confounding and maximize correlation.
\item We demonstrate that DEBIAS effectively eliminates latent confounding and consistently outperforms existing algorithms in depression and schizophrenia.
\end{enumerate}
Collectively, these advances enable robust causal inference from complex longitudinal datasets by learning \textit{multiple} durable severity scores that are empirically unconfounded, even in the presence of latent sources of bias.
\end{tcolorbox}

\section{Related Work}

We introduce the term \textit{outcome learning} to describe algorithmic identification of the optimal set or combination of outcome variables for a given analytical task. Traditionally, investigators predefine a single clinically meaningful outcome for prediction \cite{Strobl24_SV,Strobl25_bup}, which is effective when the target is clear, but problematic in complex diseases like depression or psychosis where symptoms are heterogeneous and not easily summarized by a single composite measure.

Recent algorithms in precision psychiatry attempt to learn multiple outcome measures tailored to specific analytical goals. For example, the Supervised Varimax algorithm constructs outcome measures that maximally differentiate psychiatric treatments in RCTs \cite{Strobl24_SV,Strobl25_bup}, and the method has been modified to detect subtle differences between subgroups within patient populations \cite{Strobl25}. The Sparse Canonical Outcome Regression (SCORE) algorithm, in contrast, optimizes multiple outcome measures for predictability rather than discrimination between groups \cite{Strobl2025_SCORE}. Thus, different outcome learning approaches have emerged depending on whether the goal is to distinguish treatments, differentiate patient subgroups, or maximize predictability. Importantly, none of these existing methods explicitly optimize outcome measures to enable causal inference -- a gap the DEBIAS algorithm is designed to fill.

Almost all existing causal inference methods address confounding by transforming or leveraging observed covariates but leave outcome variables fixed. For example, Difference-in-Differences relies on observed covariates to support the parallel trends assumption \cite{Ashenfelter85}, and Inverse Probability of Treatment Weighting (IPTW) uses propensity scores based on observed covariates to balance treatment groups \cite{Robins00}. In principle, IPTW can be combined with outcome-learning methods such as Non-Negative Canonical Correlation Analysis (NNCCA) \cite{Sigg07}, but such strategies still rest on the assumption that adjustment for observed covariates is sufficient to remove confounding. More recent approaches -- including meta-learners\cite{Nie21,Kunzel19} and causal forests\cite{Wager18} -- extend this logic by using flexible machine learning models to control for high-dimensional covariates and estimate the conditional average treatment effect (CATE). Yet, all these methods implicitly assume that confounding can only be addressed by modifying the set of predictors, rather than the outcomes themselves.

Instrumental variable (IV) methods provide an alternative approach by leveraging external sources of variation to identify causal effects\cite{Angrist96}. However, they depend on the strong assumption of no latent confounding linking the instrument (e.g., $T_1$) to the outcome -- an assumption that is rarely plausible in complex observational settings. In practice, observed covariates and putative instruments are frequently inadequate to eliminate confounding for all outcome variables. Moreover, no empirical test can definitively establish that all sources of bias have been addressed by observed covariates\cite{Imbens15}. Our approach instead relaxes the requirement by finding projections of the outcomes that minimize confounding, thereby retaining robustness even in the presence of residual latent confounders.

In summary, outcome learning methods in psychiatry have so far focused on RCT settings, subgroup differentiation, or improving predictivity rather than optimizing for causal inference from observational data. Meanwhile, causal inference algorithms have almost exclusively targeted the covariate space, neglecting the potential of outcome transformation. To our knowledge, DEBIAS is the first method to empirically learn outcome definitions specifically for causal inference within a unified, testable, and interpretable framework.

\section{Assumptions} \label{sec:assump}
We adopt the potential outcomes framework to rigorously formalize causal inference in longitudinal data. While Figure \ref{fig:main} illustrates the case of two treatment time points ($T_1$ and $T_2$) for simplicity, our general framework allows the treatment of interest $T_p$ to occur at any time step $p$, with the possibility of multiple prior treatments $T_j$ for all $j < p$. Moreover, to unambiguously distinguish between temporal indices and hypothetical interventions in the outcomes, we denote by $\bm{Y}_i(t_p)$ the observed or potential outcome vector at time point $i$, under the hypothetical scenario in which treatment $T_p = t_p$ was assigned at time $p < i$. This notation enables a clear analysis of intervention effects across time.

Our methodology is predicated upon three standard assumptions from the causal inference literature \cite{Imbens15}:
\begin{enumerate}[leftmargin=*]
    \item \textbf{Consistency}: If a unit receives treatment $t_p$, then the observed outcome at time $i$ coincides with the potential outcome under $t_p$; that is, $\bm{Y}_i = \bm{Y}_i(t_p)$ when $T_p = t_p$.
    \item \textbf{Stable Unit Treatment Value Assumption (SUTVA)}: The potential outcomes for any unit are unaffected by the treatment assignments of other units (i.e., there is no interference), and each treatment condition is consistently and uniquely defined.
    \item \textbf{Positivity}: The conditional density of treatment assignment is strictly positive; that is, $p(T_p = t_p \mid T_1, \bm{X}) > 0$ for all $(T_1, \bm{X}, t_p)$ within the support of the data. Similarly, $p(T_j = t_j \mid T_p, \bm{X}) > 0$ for all $j < p$ and all $(T_p, \bm{X}, t_j)$ in the support.
\end{enumerate}

Most existing studies also assume \textit{unconfoundedness} (ignorability), where $\bm{Y}_i(t_p)\alpha \ci T_p | \bm{X}$ for some fixed weight vector $\alpha \geq 0$ and all $t_p$. For example, $\bm{Y}_i(t_p)\alpha$ with $\alpha = 1$ corresponds to the total score at time point $i$ that would be observed if $T_p = t_p$. However, in practice, a potential outcome $\bm{Y}_i(t_p)\alpha$ may remain probabilistically dependent on $T_p$ given $\bm{X}$ due to confounding from latent variables like $\bm{C}$ (Figure \ref{fig:main} (a)) \cite{Richardson13}. The dependence can also extend to earlier treatments $T_j$ for all $j < p$ because the factors influencing earlier treatment assignments frequently continue to affect the assignment of treatment at the time point of interest. 

We forego the standard unconfoundedness assumption and instead propose to \textit{learn} $\alpha$ such that the antecedent in the following assumption holds:
\begin{enumerate}[leftmargin=*]
\setcounter{enumi}{3}
    \item \textbf{Projected Unconfoundedness}: If there exists $\alpha \geq 0$ such that $\bm{Y}_i(T_p)\alpha \ci \{T_1, \dots, T_{p-1} \} | T_p \cup \bm{X}$  for all $i > p$, then $\bm{Y}_i(t_p)\alpha \ci T_p | \bm{X}$ for all $i > p$ and all $t_p$.
\end{enumerate}
This condition is justified by the temporal and graphical structure of confounding in longitudinal data. Specifically, latent factors $\bm{C}$ -- such as historical severity or health insurance status -- often exert influence on both historical treatments ${T_1, \ldots, T_{p-1}}$ and the treatment of interest $T_p$, as depicted in Figure \ref{fig:main} (a). Moreover, historical treatments frequently have a direct causal effect on $T_p$ but are time-limited without direct effects on $\bm{Y}_i$ for any $i > p$. This induces a graphical structure where $T_p$ functions as a collider on backdoor paths of the form $T_j \rightarrow T_p \leftarrow \bm{C} \rightarrow \bm{Y}_i\alpha$ for some $j < p$. By choosing $\alpha$ such that $\bm{Y}_i(T_p)\alpha \ci \{T_1, \dots, T_{p-1} \} | T_p \cup \bm{X}$, we eliminate statistical dependence induced by shared latent confounders, thereby removing bias from unmeasured confounding along non-causal paths (Figure \ref{fig:main} (b)). Consequently, the standard unconfoundedness condition is restored for the specific learned outcome $\bm{Y}_i\alpha$, enabling unbiased causal effect estimation of $T_p$ on $\bm{Y}_i \alpha$ for each time point $i > p$. See Supplementary Materials \ref{SM:relax} for an even more general treatment that requires time-limited direct effects for only a subset of items.

Notice that our outcome-centric strategy departs from conventional approaches that presume no unmeasured confounding for combinations of all outcome items, such as total severity scores or remission rates -- a strong and generally untestable assumption. Our approach algorithmically learns a non-negative weight vector $\alpha$ that is typically sparse in practice, so the resulting outcome score depends only on a small subset of items. Accordingly, the projected unconfoundedness condition and resulting causal guarantees apply specifically to this empirically identified subset of outcomes, rather than to composite measures constructed a priori without regard to confounding structure. This yields both improved interpretability and greater robustness to latent confounding, as the learned outcome focuses on those items for which credible causal inference is feasible.

\section{Maximizing Correlation to Causally Distinguish Patients} \label{sec:corr}

\begin{figure}[t]
    \centering
    \includegraphics[width=1\linewidth]{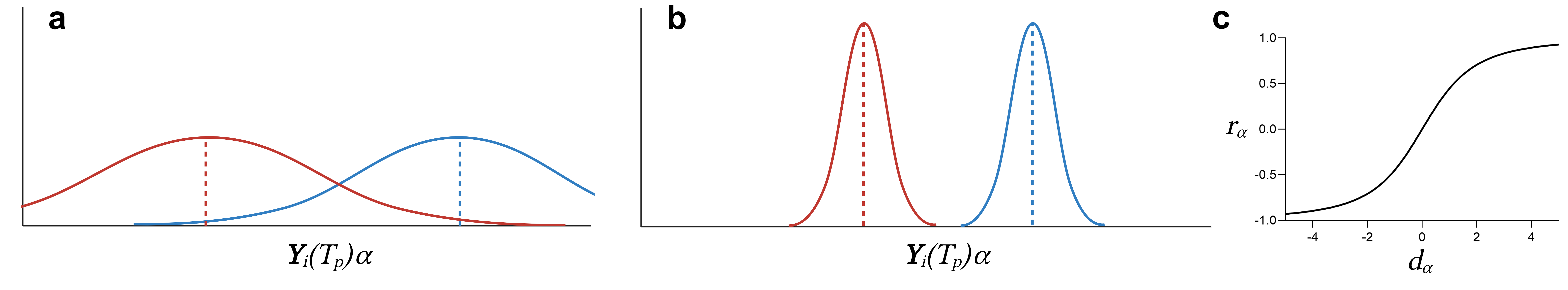} \vspace{-5mm}
    \caption{ \textbf{Comparison of effect size measures.} (a) The ATE quantifies the raw difference in group means (red vs. blue), without accounting for within-group variability or distributional overlap. (b) Cohen’s $d$ and Pearson’s correlation coefficient both capture the separability of groups by standardizing the mean difference using the pooled within-group standard deviation; separability increases as within-group variance decreases and group means diverge. (c) Although Cohen’s $d$ and correlation are monotonically related under binary treatment, only correlation extends naturally to ordinal (more than two levels) or continuous treatment variables, making it more broadly applicable. }
    \label{fig:mean_dif} 
\end{figure}

In causal inference, larger values of $\bm{Y}_i(T_p)$ are conventionally interpreted as improvement, whereas most clinical rating scales assign higher scores to greater symptom severity. To align these conventions, we multiply each rating item by $-1$ so that higher values indicate improvement. The canonical quantity for binary treatment assignment with an outcome composite is then the average treatment effect (ATE):
\begin{equation*}
     \mathrm{ATE}_{\alpha} = \mathbb{E}[\bm{Y}_i(1)\alpha] - \mathbb{E}[\bm{Y}_i(0)\alpha],
\end{equation*}
where $\bm{Y}_i(1)\alpha$ and $\bm{Y}_i(0)\alpha$ denote the potential outcomes under treatment and control, projected via $\alpha \geq 0$. The goal is to identify $\alpha$ that maximizes the treatment effect along a learned score. However, maximizing the ATE alone does not account for the variance of $\bm{Y}_i(T_p)\alpha$ within each treatment group. As a result, even if the group means are well separated, the individual outcomes may still overlap substantially due to high within-group variance (Figure \ref{fig:mean_dif} (a)), rendering it difficult to predict which patients will benefit from treatment in practice.

The separation between treatment assignments must be meaningful relative to the within-group variability of outcomes. Let $p = P(T = 1)$ denote the proportion of individuals assigned to the treatment group, and $1 - p = P(T = 0)$ the proportion in the control group. One classical approach to quantifying such standardized separation is to maximize Cohen’s $d$:
\begin{equation*} 
d_{\alpha} = \frac{\mathrm{ATE}_{\alpha}}{\sqrt{p \textnormal{Var}(\bm{Y}_i(1)\alpha) + (1-p) \textnormal{Var}(\bm{Y}_i(0)\alpha)}},
\end{equation*}
where the denominator represents the pooled within-group variance \cite{Cohen13}. Cohen’s $d$ thus quantifies the magnitude of group separation relative to internal variability. Maximizing $d$ aligns the learned outcome score with maximal distinguishability of treatment groups (Figure \ref{fig:mean_dif} (b)).

Cohen’s $d$ is unfortunately limited to binary treatment assignment, but the measure admits a monotonic transformation into Pearson’s correlation coefficient\cite{Lipsey01}:
\begin{equation*}
r_{\alpha} = \frac{\sqrt{p(1-p)} \cdot d_\alpha}{\sqrt{1 + p(1-p) \cdot d_\alpha^2}}.
\end{equation*}
This is a bijective mapping that preserves the ranking of outcome scores (Figure \ref{fig:mean_dif} (c)), so maximizing $d_\alpha$ necessarily maximizes $r_\alpha$ and vice versa. Importantly, while Cohen’s $d$ is limited to binary comparisons, Pearson’s correlation generalizes naturally to ordinal and continuous treatment variables (e.g., income levels or medication dosages), thereby offering a more versatile and broadly applicable objective. Accordingly, we adopt the correlation coefficient as our primary optimization criterion, given its capacity to maximally distinguish between treatment conditions regardless of whether the treatment variable is discrete or continuous.

\section{Algorithm} \label{sec:alg}

\subsection{First Learned Severity Score}

We now describe the DEBIAS algorithm. The algorithm begins by constructing the first non-negatively weighted outcome score that is durably responsive to treatment and robust to confounding. As detailed in Section~\ref{sec:corr}, our objective is to maximize the correlation between $T_p$ and each outcome measure $\bm{Y}_i\alpha$ for $i > p$. This leads to the following optimization problem:
\begin{equation} \label{eq:first}
\begin{aligned}
\argmax_{\alpha \geq 0, \|\alpha\|_1=1} \sum_{i=p+1}^{m} \Big[\underbrace{\textnormal{cor}(\bm{Y}_i \alpha, T_p|T_1,\bm{X})}_{(a)} - \underbrace{\frac{\lambda}{p-1}\sum_{j=1}^{p-1}\textnormal{cor}^2(\bm{Y}_i \alpha, T_j|T_p,\bm{X})}_{(b)} \Big],
\end{aligned}
\end{equation}
where we enforce $\|\alpha\|_1 = 1$ to ensure uniqueness of the solution.

The \textbf{main correlation} term in (a) seeks to maximize the durable partial correlation between the learned outcome and the target treatment across all future time points, adjusting for $\bm{X}$ to account for observed confounders. We also adjust for the most distant prior treatment $T_1$ to ensure the algorithm remains robust when treatment assignment is constant over time; without this step, the confounding penalty in (b) -- explained below -- becomes vacuously zero and cannot control for confounding. Including $T_1$ as a covariate in (a) ensures the correlation objective is also zero in such cases, preventing spurious associations when there is no temporal variation in treatment.

The \textbf{confounding penalty} in (b) targets association arising from latent confounding. Here, $\sum_{j=1}^{p-1}\textnormal{cor}^2(\bm{Y}_i \alpha, T_j|T_p, \bm{X}) = 0$ is a necessary condition for $\bm{Y}_i(T_p)\alpha \ci \{T_1, \dots, T_{p-1} \} | T_p \cup \bm{X}$ in the projected unconfoundedness condition (Section \ref{sec:assump}). Minimizing this penalty thus addresses confounding and also serves as a regularizer in finite samples: if the empirical squared partial correlation is large, it counteracts over-optimistic correlations in (a). Notably, this approach uses a linear approximation to conditional independence, enabling empirical unconfoundedness and statistical hypothesis testing of latent confounding; while not equivalent to full conditional independence, partial decorrelation is often sufficient for practical purposes in high-dimensional observational data.

To balance predictive accuracy and robustness, we optimize $\lambda$ by cross-validation over a predefined grid and select the value that maximizes the correlation in (a), \textit{provided} that the p-value associated with the squared correlation in (b) exceeds a specified threshold (default $0.05$), meaning we cannot reject the null hypothesis of no partial association. In this way, the penalty both mitigates confounding and regularizes the model to avoid overfitting.

\subsection{Sequential Extraction of Severity Scores}

The optimization problem in Expression \eqref{eq:first} only produces a single set of non-negative weights, $\alpha_1$. However, our goal is to identify all $\alpha$ that maximize persistent correlation with treatment while minimizing confounding. We thus introduce an additional \textbf{diversity-promoting penalty} term (c) that encourages each new score to be as clinically distinct as possible from all previously extracted scores:

\begin{equation} \label{eq:seq}
\begin{aligned}
\argmax_{\alpha \geq 0, \|\alpha\|_1=1} \sum_{i=p+1}^{m} \Bigg[ &\textnormal{cor}(\bm{Y}_i \alpha, T_p|T_1,\bm{X}) -  \frac{\lambda}{p-1}\sum_{j=1}^{p-1}\textnormal{cor}^2(\bm{Y}_i \alpha, T_j|T_p,\bm{X}) \\ &- \underbrace{\frac{1}{K-1}\sum_{k=1}^{K-1} \left( \frac{\alpha_k^T M_i \alpha}{\sqrt{\alpha_k^T M_i \alpha_k}\sqrt{\alpha^T M_i \alpha}} \right)}_{(c)} \Bigg],
\end{aligned}
\end{equation}
where $K \geq 2$ denotes the total number of learned scores including the current score under optimization. The diversity-promoting penalty measures the Mahalanobis cosine similarity between the current score $\alpha$ and each previously discovered score $\alpha_k$, where $M_i$ is the correlation matrix of the outcome variables $\bm{Y}_i$. This similarity ranges from $-1$ to $1$ like the main correlation term, so the penalty is naturally scale-matched to the main objective and does not require a separate hyperparameter for balancing its influence.

We intentionally do \textit{not} square the Mahalanobis cosine similarity in the diversity-promoting penalty so that the optimization procedure favors negative similarity values when permitted by the data and the non-negativity constraint. Recall that (1) the individual items in $\bm{Y}_i$ represent \textit{pathological} symptoms, where higher values indicate worse clinical severity, and (2) each item was sign-flipped in Section \ref{sec:corr} so that higher values instead denote clinical \textit{improvement}, aligning with the causal interpretation of treatment effects. Within this transformed space, all weights in $\alpha$ are constrained to be non-negative, ensuring that learned outcome scores reflect additive improvements across symptoms without cancellation. These two conditions impose a directional structure: outcome scores represent strictly non-negative combinations of improvements in underlying pathologies. Consequently, negative Mahalanobis cosine similarity between scores becomes possible only when the symptom correlation matrix $M_i$ contains sufficiently negative off-diagonal elements -- i.e., when some symptoms improve inversely with others. In such cases, the algorithm can extract new scores that are anti-aligned with prior ones within the non-negative feasible region, capturing distinct, clinically interpretable axes of symptom variation.

For example, manic and depressive symptoms in bipolar disorder are clearly clinically distinct yet strongly negatively correlated. In this context, DEBIAS may recover one composite score weighted toward manic symptoms and another toward depressive symptoms, both with non-negative weights and exhibiting negative Mahalanobis cosine similarity. In contrast, minimizing the squared Mahalanobis cosine similarity only enforces orthogonality, which can result in a new score that emphasizes peripheral symptoms -- such as chronic irritability, distractibility, or somatic complaints -- rather than capturing the strong negative correlation between core mood domains. The diversity-promoting penalty thus directly exploits the negative correlation structure to extract outcome scores that are clinically distinct, rather than conflating mathematical orthogonality with clinical distinctiveness.

We ultimately solve the optimization problem in Expressions \eqref{eq:first} and \eqref{eq:seq} by projected gradient ascent using backtracking line search with the Armijo condition \cite{Armijo66} (Supplementary Materials \ref{SM:pseudocode}). Computational complexity analysis revealed that DEBIAS scales linearly with the number of subjects and time points, but quadratically with the number of outcome items, covariates, and summary scores (Supplementary Materials \ref{SM:complexity}).

\section{Results}

\subsection{Comparator Algorithms}

We ran DEBIAS with 5-fold cross-validation evaluating $\lambda$ over the grid ${0, 1, \ldots, 10}$. We extracted $s=3$ severity scores. We compared DEBIAS against the following algorithms:
\begin{enumerate}[leftmargin=*]
\item \textbf{Non-Negative Canonical Correlation Analysis (NNCCA) with IPTW}\cite{Sigg07}: NNCCA estimates time point-specific outcome weights by independently maximizing the IPTW-weighted correlation at each time point: $\max_{\alpha \geq 0} \mathrm{cor}_w(\bm{Y}_i \alpha, T_p)$. NNCCA further applies a weighted deflation procedure to iteratively extract approximately orthogonal sets of non-negative outcome weights.
\item \textbf{R-Learner with Extreme Gradient Boosting (RBoost)}\cite{Nie21}: RBoost first estimates the baseline outcome function and the propensity score using XGBoost\cite{Chen16}, and then computes residualized outcomes and residualized treatment assignments. Next, the algorithm fits a model for the CATE by regressing the residualized outcomes onto the residualized treatments using XGBoost. This orthogonalization-based procedure improves robustness to model misspecification and enables efficient use of modern machine learning methods for CATE estimation. We used 3 folds for cross-fitting and cross-validation, 100 trees, 3 search rounds, and 5 early stopping rounds.

\item \textbf{Causal Forests (CF)}\cite{Wager18}: Causal forests estimate the CATE by building an ensemble of honest causal trees, each of which partitions the feature space to locally estimate treatment effects. Honesty is ensured by splitting the data into separate subsamples for tree construction and effect estimation. Averaging over the ensemble yields a flexible, nonparametric estimator with valid asymptotic inference. We learned 2000 trees and otherwise used default parameters.
\end{enumerate}
We further assessed DEBIAS against \textbf{three ablated variants}: (i) replacing the correlation objective with mean squared error (MSE), (ii) removing the confounding penalty in Expression \eqref{eq:seq} ($\lambda = 0$), and (iii) replacing the correlation objective and removing the confounding penalty. Notably, the variant with the confounding penalty removed can be considered a variant of the SCORE algorithm \cite{Strobl2025_SCORE}, except that we employ a diversity-promoting penalty (Mahalanobis cosine similarity) instead of deflation to extract multiple scores when treatment is univariate (a setting where deflation fails). All ablated variants used the same cross-validation protocol, $\lambda$ grid, and number of scores as DEBIAS.

DEBIAS differs from existing comparator methods along several key dimensions. First, while all comparators adjust only for observed confounding, DEBIAS uniquely leverages the temporal structure of longitudinal data and the time-limited nature of treatments to additionally reduce the influence of latent confounding. Second, DEBIAS is the only method capable of extracting multiple interpretable severity scores in the setting of binary treatment assignment, owing to its use of a diversity-promoting penalty rather than deflation. Although NNCCA can also produce multiple scores, these do not correspond to strict severity scores and lack a clear causal interpretation. Third, DEBIAS estimates non-negative outcome weights by jointly utilizing all future time points, thereby borrowing statistical strength across the outcome trajectory. In contrast, NNCCA, RBoost, and CF estimate effects or representations independently at each time point, precluding information sharing across time. Taken together, DEBIAS is the only approach that integrates temporal consistency and confounding control -- both observed and latent -- to derive outcome scores that are both longitudinally predictive and causally valid.

\subsection{Evaluation Metrics}

As detailed in Section~\ref{sec:corr}, our primary goal is to assess how effectively each method differentiates patients based on treatment assignment, motivating the use of \textbf{partial correlation} $\mathrm{cor}(\widehat{m}_i, T_2 \mid T_1, \bm{X})$ as the main performance metric for each time point $i > 2$; here, $\widehat{m}_i$ represents the learned outcome at time $i$ -- specifically, $\bm{Y}_i \alpha$ for DEBIAS and NNCCA, or the estimated CATE of the total severity score for RBoost and CF. However, naive correlation is only meaningful in the absence of confounding. To address this limitation, we additionally evaluate all algorithms using the \textbf{confounding p-value} derived from the squared partial correlation, $\mathrm{cor}^2(\widehat{m}_i, T_1 \mid T_2, \bm{X})$. This squared partial correlation quantifies the residual association between the model output and historical exposures, reflecting the extent of any remaining confounding bias. We further evaluated each method by the \textbf{sum of confounded coefficients}, or the sum of the coefficients associated with outcome items known to suffer from latent confounding; methods that achieve a low total are preferred, as they minimize reliance on confounded information. Finally, we compare all algorithms in terms of \textbf{run-time} to evaluate computational efficiency.

We evaluated all metrics using 1,000 bootstrap replicates, where we sampled with replacement such that approximately 63.2\% of the samples acted as the training set and the remaining 36.8\% as the held-out test set in each replicate. We computed all performance measures exclusively on the test samples. For all comparisons of relative algorithm performance, we applied two-sided paired $t$-tests and interpreted significance at a Bonferroni-corrected threshold of $0.05/u$, where $u$ denotes the total number of comparator algorithms.

\subsection{Adolescent Depression}

We evaluated each algorithm's ability to maximize correlation and minimize confounding using data from the Treatment for Adolescents with Depression Study (TADS; ClinicalTrials.gov ID NCT00006286), an RCT assessing the efficacy of cognitive-behavioral therapy (CBT), fluoxetine, their combination, and placebo in adolescent major depressive disorder \cite{March04}. We excluded the placebo group since patients received placebo for only 8 weeks before transitioning to standard care, whereas 323 patients in the other groups maintained their assigned treatments for a 36-week duration.

NNCCA, CF, and RBoost are canonically applied to binary treatment variables; therefore, treatment groups were dichotomized as medication (fluoxetine or combination therapy) versus CBT to facilitate a fair comparison across methods. The resulting medication indicator was used as the current treatment variable, $T_p = T_2$. The prior treatment indicator, $T_1$, was defined as a binary variable indicating whether the patient had received any mental health services prior to treatment assignment, as measured by the Child and Adolescent Services Assessment (CASA)\cite{Ascher96}, given its likely correlation with $T_2$. The outcome vector $\bm{Y}_i$ comprised the change from baseline in each item of the Children’s Depression Rating Scale--Revised (CDRS-R)\cite{Poznanski95}.

Although the original RCT design ensures perfect control of confounding, we introduced an artificial latent confounder, $\bm{C}$, sampled from a standard normal distribution. We then additively incorporated $\bm{C}$ into $T_1$, $T_2$, and a randomly selected subset of 5 to 12 items within each outcome vector $\bm{Y}_i$, using time-varying weights for each item and time point sampled uniformly from $[0.2, 1]$. We then re-binarized $T_1$ and $T_2$ by thresholding at the value 0.25. Crucially, this experimental setup provides ground-truth knowledge regarding which $\bm{Y}_i$ items are confounded to compute the sum of confounded coefficients metric. All analyses included age and biological sex as observed covariates in $\bm{X}$.

\begin{figure}[t]
    \centering
    \includegraphics[width=1\linewidth]{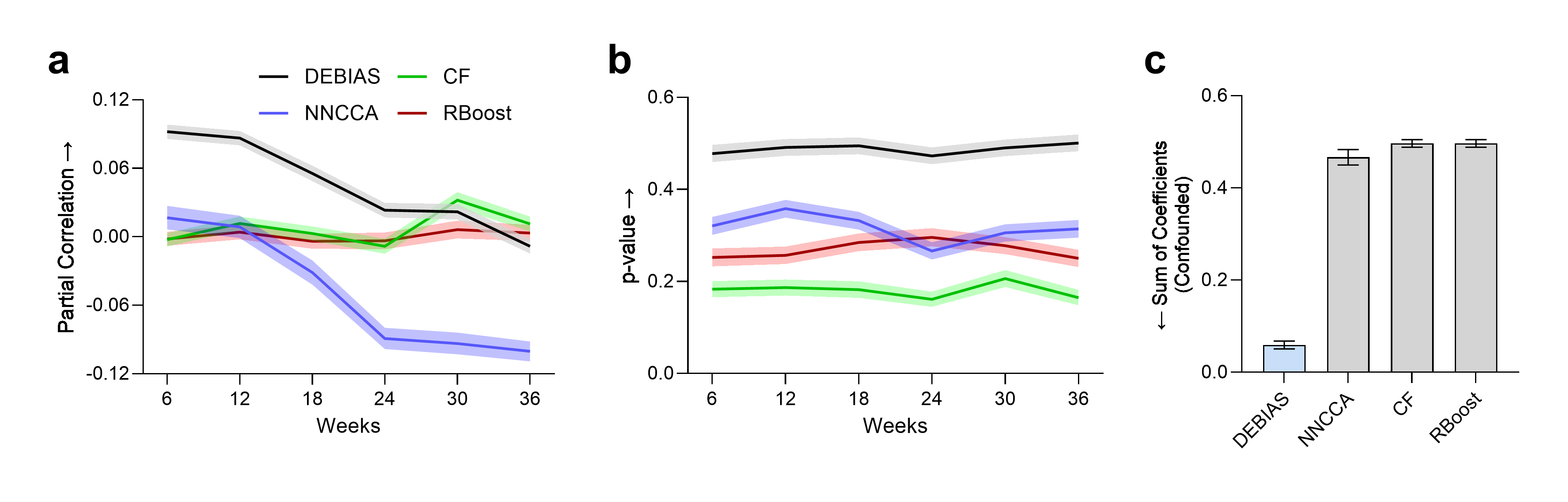} \vspace{-9mm}
    \caption{\textbf{Accuracy results for the first score in the TADS dataset.} (a) DEBIAS achieved the highest correlation with outcomes from weeks 6-24, despite incorporating regularization to mitigate latent confounding bias. (b) DEBIAS also attained the largest mean p-value associated with the squared correlation coefficient, reflecting reduced confounding influence relative to competing methods. (c) The sum of the outcome coefficients assigned by DEBIAS to confounded items was also the smallest among all methods. Error bands and bars denote 95\% confidence intervals of the mean.}
    \label{fig:TADS}
\end{figure}

Figure \ref{fig:TADS} presents the primary results. DEBIAS achieved the highest correlation coefficients among all methods across the first 24 weeks (Figure \ref{fig:TADS} (a)). This superior performance reflects the enforcement of a consistent set of non-negative weights $\alpha$, which facilitates information sharing across time points. In later weeks, all algorithms exhibited a decline toward zero or negative correlations, reflecting the true convergence of treatment effects among fluoxetine, CBT, and combination therapy\cite{March04}. We conclude that DEBIAS maximizes outcome predictability relative to all comparators.

\begin{figure}
    \centering
    \includegraphics[width=0.37\linewidth]{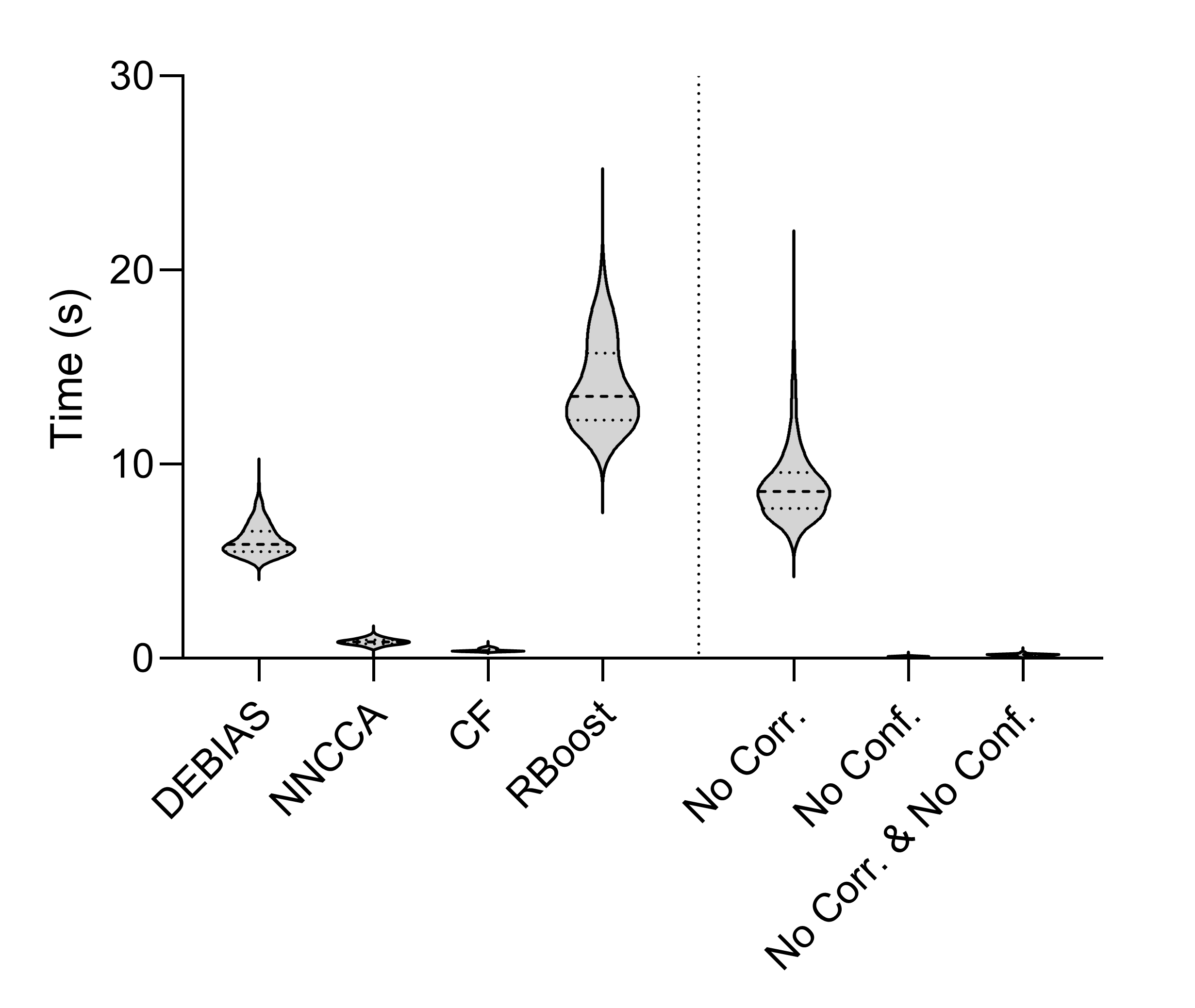}
    \caption{\textbf{Timing results for TADS.} Violin plots showing that DEBIAS usually completed within 10 seconds, slightly faster than RBoost. Dotted lines within each violin denote the 25th, 50th and 75th percentiles.}
    \label{fig:TADS:time}
\end{figure}

DEBIAS also yielded the largest p-values associated with the squared correlation regularization term, indicating effective mitigation of latent confounding effects (Figure \ref{fig:TADS} (b)). Inspection of the outcome coefficients revealed that DEBIAS preferentially assigned the smallest weights to the confounded outcomes, providing further evidence of superior confounding control (Figure \ref{fig:TADS} (c)). Collectively, these results demonstrate that DEBIAS achieves both maximal correlation and minimal confounding in the TADS dataset. Ablation results against three variants in Supplementary Figure \ref{fig:TADS:ablation} confirmed that both correlation maximization and confounding minimization are essential for DEBIAS to achieve optimal performance, even for the scores beyond the first. Finally, DEBIAS completed within 10 seconds on average in this dataset (Supplementary Figure \ref{fig:TADS:time}).

\subsection{Chronic Schizophrenia}

\begin{figure}[b]
    \centering
    \includegraphics[width=1\linewidth]{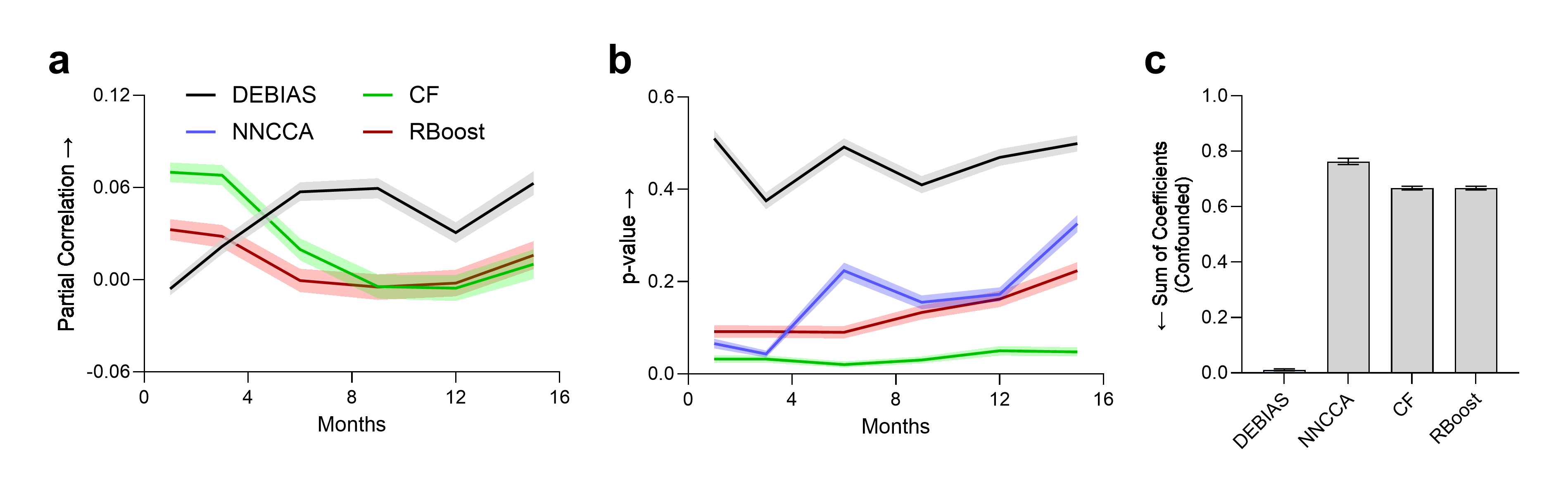} \vspace{-9mm}
    \caption{\textbf{Accuracy results for the first score in the CATIE dataset.} (a) DEBIAS attained the highest correlation with treatment assignment as treatment effects diverged over time. The correlation achieved by NNCCA was consistently poor (below -0.20 on average) and therefore omitted from the plot to focus on the competitive algorithms. (b) Compared to all other methods, DEBIAS exhibited substantially superior control of confounding, as reflected by higher p-values for the squared correlation coefficient. (c) DEBIAS also assigned markedly lower non-zero coefficients to confounded outcome items, indicating greater selectivity and robustness to confounding.}
    \label{fig:CATIE}
\end{figure}

We next evaluated each algorithm's ability to learn causally predictable outcomes in schizophrenia. We downloaded data from the Clinical Antipsychotic Trials of Intervention Effectiveness (CATIE; NCT00014001), which was a large RCT comparing the effectiveness of antipsychotic medications in adults with chronic schizophrenia \cite{Lieberman05}. We restricted our analysis to 664 subjects treated with olanzapine (coded as $T_2 = T_p = 1$) or quetiapine ($T_2=0$), in order to maximize the contrast in antipsychotic efficacy for core psychotic symptoms; olanzapine is widely regarded as one of the most effective antipsychotics, whereas quetiapine is generally considered among the least effective \cite{Leucht13}. We set $T_1$ to the number of prior inpatient psychiatric visits, since this variable influences antipsychotic prescribing in patients with schizophrenia \cite{Tiihonen17}. We introduced time-varying confounding by additively including a standard normal latent confounder in $T_1$, $T_2$, and a uniformly selected subset of 15 to 25 items within each outcome vector $\bm{Y}_i$. Each outcome vector represented the change from baseline for the 30 individual items of the Positive and Negative Syndrome Scale (PANSS) measured across 15 months \cite{Kay87}.

\begin{figure}
    \centering
    \includegraphics[width=1\linewidth]{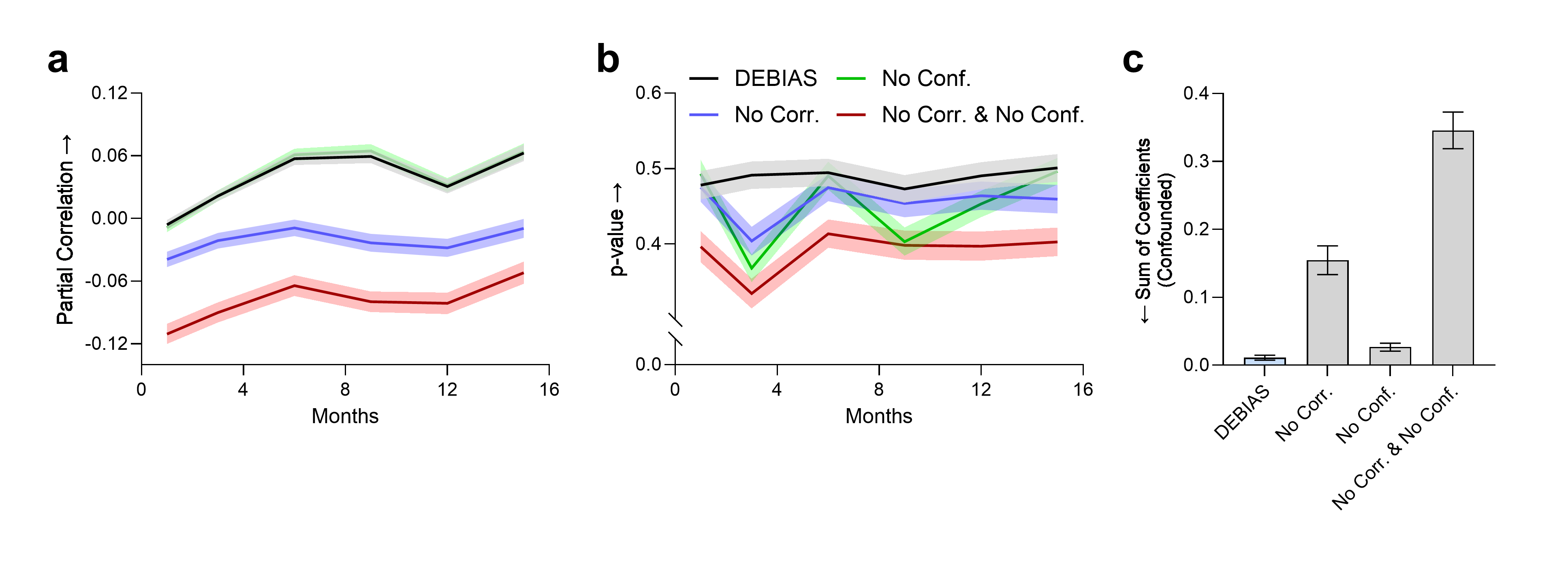}
    \caption{\textbf{Ablation accuracy results for CATIE.} For the first recovered score, both DEBIAS and the variant without the confounding penalty (no conf.) achieved the highest correlation with treatment assignment in (a). However, DEBIAS demonstrated superior control of confounding, as indicated by the largest p-values for the squared correlation coefficient in (b) and the lowest sum of coefficients assigned to confounded outcome items in (c). For the second and third scores, all methods produced similarly low, near-zero correlations; these results are therefore not shown.}
    \label{fig:CATIE:ablation}
\end{figure}

\begin{figure}
    \centering
    \includegraphics[width=0.37\linewidth]{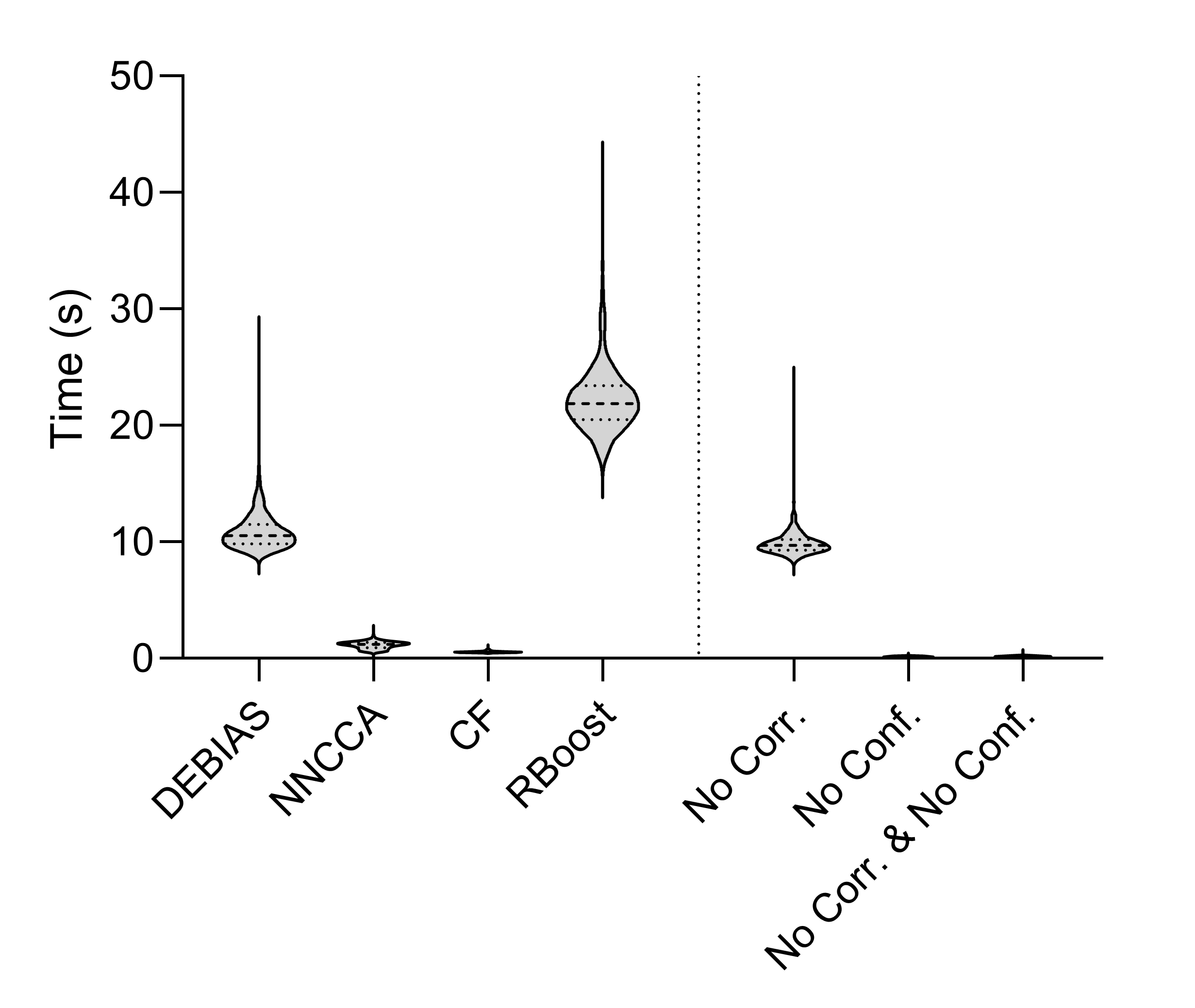}
    \caption{\textbf{Timing results for CATIE.} DEBIAS completed within approximately 20 seconds on this dataset, outperforming RBoost in computational efficiency but requiring more time than the remaining comparator algorithms.}
    \label{fig:CATIE:time}
\end{figure}

We present the results in Figure \ref{fig:CATIE}. DEBIAS once again achieved the highest correlation with treatment assignment among all competing methods (Figure \ref{fig:CATIE} (a)). The algorithm also improved confounding control, as evidenced by a markedly higher p-value for the confounding penalty and a much lower total sum of non-negative coefficients assigned to confounded outcome items. These results indicate that DEBIAS consistently maximized out-of-sample correlation and minimized confounding, paralleling its performance in the TADS dataset. Ablation analyses further demonstrated that both the correlation maximization objective and the confounding penalty were necessary to achieve optimal results across multiple extracted scores (Supplementary Figure \ref{fig:CATIE:ablation}). Finally, DEBIAS completed its computation within 20 seconds on this dataset (Supplementary Figure \ref{fig:CATIE:time}).

\section{Discussion}

We introduced the DEBIAS algorithm, which learns combinations of symptom severity items that maximize predictability and minimize latent confounding bias, enabling robust causal inference from psychiatric longitudinal data. Unlike conventional approaches that rely on fixed, user-defined outcomes and assume adequate adjustment for confounding, DEBIAS is the first method to algorithmically identify outcome measures empirically amenable to valid causal inference even with latent confounders. Experiments in adolescent depression and chronic schizophrenia demonstrated that DEBIAS recovers causally predictable severity scores by leveraging the time-limited direct effects of past treatments, a critical advantage when comprehensive covariate collection is infeasible. DEBIAS thus represents a paradigm shift, showing that identifying suitable outcomes for causal inference can be just as important as adjusting for observed confounders in complex real-world settings.

Our work has several limitations. First, DEBIAS is purposefully designed as a linear method to emphasize foundational conceptual advances in causal inference rather than algorithmic complexity. While linear decorrelation is often practically effective for removing confounding, it does not guarantee conditional independence in the strict mathematical sense. Thus, extending DEBIAS to nonlinear settings, where more general forms of dependence can be addressed, remains important for future research. Second, we currently treat $T_1$ as a historical treatment with time-limited direct effects, an assumption justified for most psychiatric medications. The assumption may be violated for interventions such as CBT, which is believed to exert long-lasting effects by imparting enduring skills \cite{Hollon06}. However, in practice, the presence of persistent symptoms among non-remitting patients included in clinical longitudinal datasets suggests that any enduring direct effects of prior therapy are likely minimal in this cohort. Our experimental results further support this claim: DEBIAS consistently achieved the lowest levels of confounding across all methods, despite not explicitly manipulating or blocking potential pathways of the form $T_1 \rightarrow \bm{Y}_3, \ldots, \bm{Y}_m$. This empirical finding suggests that the time-limited direct effects assumption and corresponding regularization strategy are reasonable and effective. Time-limited direct effects on all outcome items are also not required for our identification strategy, as discussed in Supplementary Materials \ref{SM:relax}. Third, outcome learning opens avenues for novel approaches to confounding adjustment and offers potential applications in domains beyond psychiatry, such as the construction of composite endpoints in endocrinology or the definition of multi-omic phenotypes in oncology. Systematically exploring these alternatives represents a particularly promising avenue for future research.

In summary, DEBIAS represents a substantial advance in outcome learning for causal inference, delivering clinically interpretable measures from data readily available in psychiatric research. This enables more accurate and unbiased causal inference for important real-world applications that may ultimately improve mental healthcare.

\bibliographystyle{ws-procs11x85}
\bibliography{biblio}

\newpage
\section{Supplementary Materials}

\subsection{Relaxing Time-Limited Direct Effects} \label{SM:relax}

\begin{figure}
    \centering
    \includegraphics[width=0.8\linewidth]{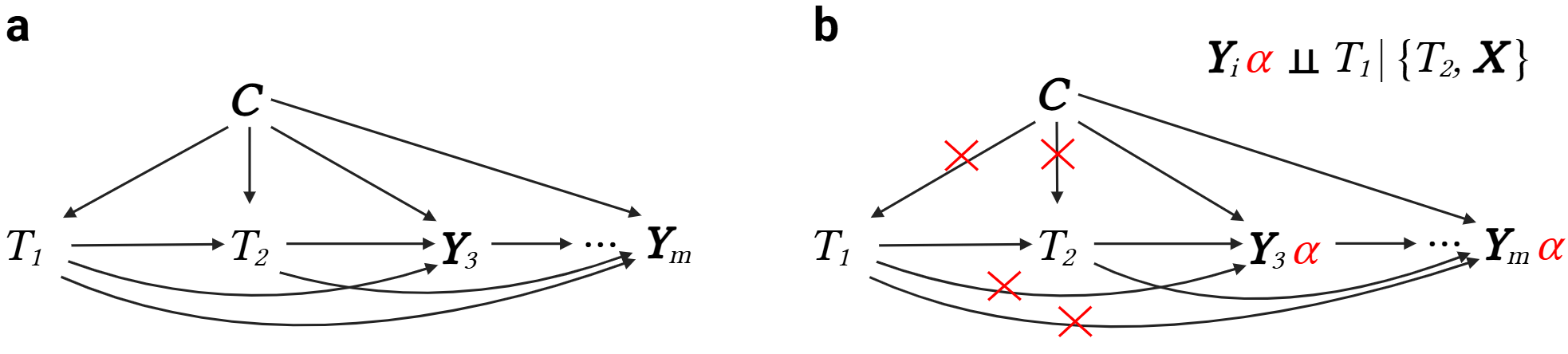}
    \caption{\textbf{Time-limited direct effects on a subset of outcomes.} For clarity, we omit the observed confounders $\bm{X}$ from the figure. (a) $T_1$ has time-limited direct effects on only a subset of items in each $\bm{Y}_i$; for many other items, $T_1$ exerts its influence indirectly, mediated through $T_2$. (b) We learn non-negative weights $\alpha \geq 0$ that eliminate statistical influence from both direct causal and non-causal paths from past treatments, e.g., as indicated by the red crosses.}
    \label{fig:relax}
\end{figure}

We can relax the assumption that time-limited direct effects apply to all items in each $\bm{Y}_i$, allowing them to instead affect only a subset of items. In Figure~\ref{fig:relax} (a), additional directed edges from $T_1$ to each $\bm{Y}_i$ represent scenarios where $T_1$ has time-limited direct effects on only a subset of the items in $\bm{Y}_i$; the effect of $T_1$ on $\bm{Y}_i$ for $i > 2$ operates exclusively through $T_2$ for the other items. Specifying the antidepressant example in the Introduction, antidepressants may not directly affect anhedonia (a specific rating scale item) a year after discontinuation in active major depression\cite{Treadway11}, even if they can directly influence other symptoms.

Achieving conditional independence between $T_1$ and $\bm{Y}_i\alpha$ (after adjusting for $T_2$ and $\bm{X}$) accomplishes two objectives:
\begin{enumerate}[label=(\alph*),leftmargin=7.2mm]
\item We remove latent confounding between $T_1$ and each $\bm{Y}_i\alpha$, and thus between $T_2$ and each $\bm{Y}_i\alpha$ due to shared latent confounders with $T_1$.
\item We eliminate any direct causal effect of $T_1$ on each $\bm{Y}_i\alpha$.
\end{enumerate}
Graphically, this approach eliminates the statistical influence of both (a) non-causal paths and (b) direct causal paths from $T_1$ to each $\bm{Y}_i\alpha$, enabling unbiased estimation of the causal effect of $T_2$ on each $\bm{Y}_i\alpha$ (Figure~\ref{fig:relax} (b)).

We can formalize the relaxation by modifying the consistency and projected unconfoundedness assumptions in Section~\ref{sec:assump}. Let $\bm{T} = \{ T_1, \dots, T_p \}$. We update consistency as follows:
\begin{enumerate}[leftmargin=*]
    \item \textbf{Consistency}: If a unit receives treatments $\bm{t}$, then the observed outcome at time $i$ coincides with the potential outcome under $\bm{t}$; that is, $\bm{Y}_i = \bm{Y}_i(\bm{t})$ when $\bm{T} = \bm{t}$.
\end{enumerate}
Here, we no longer require that time-limited direct effects from treatments $T_j$ (for all $j < p$) apply to every item in each $\bm{Y}_i$. Instead, there may exist a sparse weight vector $\alpha$ such that, for all $t_p$, the projected outcome satisfies $\bm{Y}_i(\bm{t})\alpha = \bm{Y}_i^*(t_p)$ for all treatment histories $\bm{t}$ with $T_p = t_p$. By construction, the projected outcome $\bm{Y}_i^*(t_p)$ is directly affected only by $T_p$; any influence of prior treatments ($T_j$ for $j < p$) on $\bm{Y}_i^*(t_p)$ is fully mediated by $T_p$. As a result, earlier treatments have no direct effect on $\bm{Y}_i^*(t_p)$ once $T_p$ is specified.  We thus have the following in the factual data for the realized treatment history $\bm{T}$:
\begin{equation*}
\bm{Y}_i \alpha = \bm{Y}_i(\bm{T}) \alpha = \bm{Y}_i^*(T_p),
\end{equation*}
so that the learned outcome $\bm{Y}_i \alpha$ is solely a function of $T_p$.

We then seek to satisfy the antecedent in the following revised projected unconfoundedness assumption:
\begin{enumerate}[leftmargin=*]
\setcounter{enumi}{3}
    \item \textbf{Projected Unconfoundedness}: If there exists $\alpha \geq 0$ such that $\bm{Y}_i(\bm{T})\alpha = \bm{Y}_i^*(T_p)$ and $\bm{Y}_i^*(T_p) \ci \{T_1, \dots, T_{p-1} \} | T_p \cup \bm{X}$ for all $i > p$, then $\bm{Y}_i^*(t_p) \ci T_p | \bm{X}$ for all $i > p$ and all $t_p$.
\end{enumerate}
The first condition, $\bm{Y}_i(\bm{T})\alpha = \bm{Y}_i^* (T_p)$, ensures that the weighted outcome is only directly affected by $T_p$. The second condition states that the weighted outcome and historical treatments are independent after conditioning on $T_p$ and $\bm{X}$ -- ruling out any backdoor or direct causal paths between historical treatments and outcomes. As a result, the standard unconfoundedness condition is restored for $\bm{Y}_i^*(t_p)$ for all $t_p$, allowing unbiased estimation of the causal effect of $T_p$ on $\bm{Y}_i \alpha$ for each time point $i > p$.

With this relaxed theoretical backbone in place, all subsequent concepts and experimental procedures described from Section~\ref{sec:corr} onward in the main text continue to apply seamlessly, with the minor adjustment that we return to the notation $\bm{Y}_i(T_p)\alpha$ to denote the empirically learned, non-negatively weighted outcome at time $i$ under treatment $T_p$. This expression is equivalent to $\bm{Y}_i^*(T_p)$ as defined above in the factual data: specifically, $\bm{Y}_i^*(T_p)$ represents a projected outcome that, by construction, is directly affected only by $T_p$ and not by earlier treatments, and is realized as a linear combination $\bm{Y}_i\alpha$ for a suitably chosen data-driven weight vector $\alpha$. Consequently, all algorithms and empirical analyses involving $\bm{Y}_i\alpha$ in the main text remain valid for $\bm{Y}_i^*(T_p)$ under this generalized framework.

\subsection{Pseudocode} \label{SM:pseudocode}

Let $\mathcal{L}$ denote the objective function defined in Expression \eqref{eq:seq}. We maximize $\mathcal{L}$ for a fixed $\lambda$ using projected gradient ascent, as summarized in Algorithm~\ref{alg:DEBIAS}, with the step size determined by backtracking line search satisfying the Armijo condition. At each iteration, the empirical gradient in Line~\ref{alg:grad} is computed in matrix notation as follows, under the assumption that all variables have already been residualized with respect to the appropriate conditioning terms:
\begin{equation} \label{eq:grad}
\begin{aligned}
\nabla_\alpha \mathcal{L} = \sum_{i=p+1}^m \Bigg[
    &\underbrace{\frac{\bm{Y}_i^T T_p}{\|\bm{Y}_i\alpha\|\|T_p\|} - \frac{(\bm{Y}_i\alpha)^T T_p}{\|\bm{Y}_i\alpha\|^3\|T_p\|}\bm{Y}_i^T\bm{Y}_i\alpha}_{(a)} \\
    &- \underbrace{\frac{\lambda}{p-1}\sum_{j=1}^{p-1}
    2\,\frac{ (\bm{Y}_i\alpha)^T T_j }{ \|\bm{Y}_i\alpha\| \cdot \|T_j\| } \left( 
        \frac{\bm{Y}_i^T T_j}{\|\bm{Y}_i\alpha\|\|T_j\|} - \frac{(\bm{Y}_i\alpha)^T T_j}{\|\bm{Y}_i\alpha\|^3\|T_j\|}\bm{Y}_i^T\bm{Y}_i\alpha
    \right)}_{(b)} \\
    &- \underbrace{\frac{1}{K-1}\sum_{k=1}^{K-1}
    \left(
        \frac{M_i\alpha_k}{\sqrt{\alpha_k^T M_i \alpha_k \alpha^T M_i \alpha}} - \frac{\alpha_k^T M_i \alpha}{(\alpha^T M_i \alpha)^{3/2} \sqrt{\alpha_k^T M_i \alpha_k }} M_i\alpha \right)}_{(c)}
\Bigg]
\end{aligned}
\end{equation}
The summation over Mahalanobis cosine similarities (c) is only present when $K \geq 2$. After computing the gradient, the algorithm determines an appropriate step size via backtracking line search with the Armijo condition\cite{Armijo66} (Line~\ref{alg:line_search}), updates the iterate by moving in the direction of the gradient (Line~\ref{alg:update}), and then projects onto the feasible set by enforcing non-negativity (Line~\ref{alg:project_alpha}) and $\ell_1$-normalization (Line~\ref{alg:norm_alpha}). This procedure is repeated until convergence for each of the $s$ summary scores.

\begin{algorithm*}[t]
 \textbf{Input:} Treatment at the time of interest $T_p$; historical treatments $T_j$ for $j < p$; outcome items $\bm{Y}_i$ for $i > p$; covariates $\bm{X}$; number of summary scores $s \leq$ cardinality of $\bm{Y}_i$ \\
 \textbf{Output:}  $\alpha_k$ for all $k \leq s$
\begin{algorithmic}[1]
\For{$K \in \{1,\dots, s\}$}
    \While{$\alpha$ not converged} \label{alg:while}
            \State  $\nabla f(\alpha) \leftarrow$ compute gradient for $\alpha$ (Equation \eqref{eq:grad}) \label{alg:grad}
            \State $\eta \leftarrow$ backtracking line search with the Armijo condition\label{alg:line_search}
            \State $\alpha \leftarrow \alpha + \eta \nabla f(\alpha)$ \label{alg:update}
            \State $\alpha \leftarrow \textnormal{max}(\alpha,0)$ \label{alg:project_alpha}
            \State $\alpha \leftarrow \alpha / \| \alpha \|_1$ \label{alg:norm_alpha}
    \EndWhile
    \State $\alpha_K \leftarrow \alpha$
\EndFor
\caption{DEBIAS for fixed $\lambda$} \label{alg:DEBIAS}
    \end{algorithmic}
\end{algorithm*}

We emphasize that Algorithm~\ref{alg:DEBIAS} is executed for a fixed value of the regularization parameter $\lambda$. The optimal value of $\lambda$ is selected via cross-validation to maximize the main correlation term summed over all $s$ scores:
$$\sum_{k=1}^{s} \sum_{i=p+1}^{m} \textnormal{cor}(\bm{Y}_i \alpha_k, T_p|T_1,\bm{X}),$$
but subject to the constraint that the geometric mean of the minimum p-values for the squared correlation coefficient (the confounding test), taken across scores and cross-validation folds, remains above a user-specified threshold ($\gamma = 0.05$ by default).

If no value of $\lambda$ satisfies the confounding constraint, the algorithm can proceed in one of two ways: (1) it may abstain from producing an outcome score for the current configuration, or (2) it may select the value of $\lambda$ whose associated confounding p-value is closest to, but does not exceed, the threshold $\gamma$. While abstention is preferred in practical applications to avoid unreliable inferences, we report the solution corresponding to the $\lambda$ that yields a confounding p-value closest to $0.05$ in the experiments for fair comparison against existing algorithms. After selecting the optimal $\lambda$ via cross-validation, the final model is retrained on the full dataset using this value, and the resulting summary scores are reported.

\subsection{Computational Complexity} \label{SM:complexity}

We now present a detailed analysis of the computational complexity of the DEBIAS algorithm. Let $n$ denote an upper bound on the number of subjects with complete data across all time points, $q$ the number of outcome items per vector $\bm{Y}_i$, $r$ the number of covariates in $\bm{X}$, $m$ the number of unique time points, $p$ the index of the treatment of interest, $s$ the number of extracted summary scores, $I$ the maximum number of gradient ascent iterations, and $L$ the number of line search steps per iteration. Throughout, we assume $n \gg r$.

Residualization at each time point requires $O(n(r+1)^2)$ operations, and is performed at most $(m-p)(p-1)$ times in total. Therefore, the overall complexity for residualization is $O\big( (m-p)(p-1) n(r+1)^2 \big)$. Calculation of correlation matrices $M$ for all $m-p$ time points requires $O( (m-p) nq^2)$ operations.

We now consider the computational cost of evaluating the gradient in Equation~\eqref{eq:grad} after residualization. The main correlation term (a) requires $O(nq^2)$ time, the confounding penalty (b) requires $O( (p-1) n q^2)$, and the diversity-promoting penalty (c) requires $O(s q^2)$. Each of these must be computed for each of the $m-p$ relevant time points, yielding a total cost per gradient evaluation of:
$$O\left( (m-p) \left[\, p\, n q^2 + s q^2 \,\right] \right).$$ 
Each gradient update is augmented by a backtracking line search, which entails $L$ additional objective evaluations per iteration. Excluding preprocessing, evaluation of the objective function for a single set of weights requires $O(nq)$ for the primary correlation, $O((p-1) n q)$ for the confounding penalty, and $O(s q^2)$ for the diversity-promoting penalty. The total per-evaluation cost is thus:
$$O\left( (m-p) \left[\, (p-1)\, n q + s q^2 \,\right] \right).$$ 
and the overall per-iteration complexity is increased by a factor of $L$.

Consequently, for a fixed $\lambda$, performing $I$ gradient ascent iterations for each of $s$ summary scores results in a total complexity of:
$$O\left((m-p)(p-1) n(r+1)^2 + (m-p) nq^2 +  sI L (m-p) \left[\, (p-1)\, n q + s q^2 \right] + sI  (m-p) \left[\, p\, n q^2 + s q^2 \,\right] \right).$$ 
This can be equivalently expressed as:
$$O \left(
(m-p) \left[
(p-1)\, n\, (r+1)^2
+ n\, q^2\, (1 + s I p)
+ s I L\, (p-1)\, n\, q
+ s^2 I (L+1) q^2
\right]
\right).$$
In summary, the computational complexity of DEBIAS scales linearly with the number of subjects ($n$) and time points ($m$), but quadratically with the number of outcome items ($q$), covariates ($r$), and summary scores ($s$). In most practical settings, $m$ and $q$ remain moderate, and $s$ is typically much smaller than $q$, so the algorithm remains computationally tractable for datasets with tens of thousands of subjects and hundreds of outcome variables. When $\lambda$ is selected via cross-validation, the total computational cost increases proportionally with the number of candidate $\lambda$ values times the number of cross-validation folds.

\newpage
\subsection{Additional TADS Results}

\begin{figure}
    \centering
    \includegraphics[width=1\linewidth]{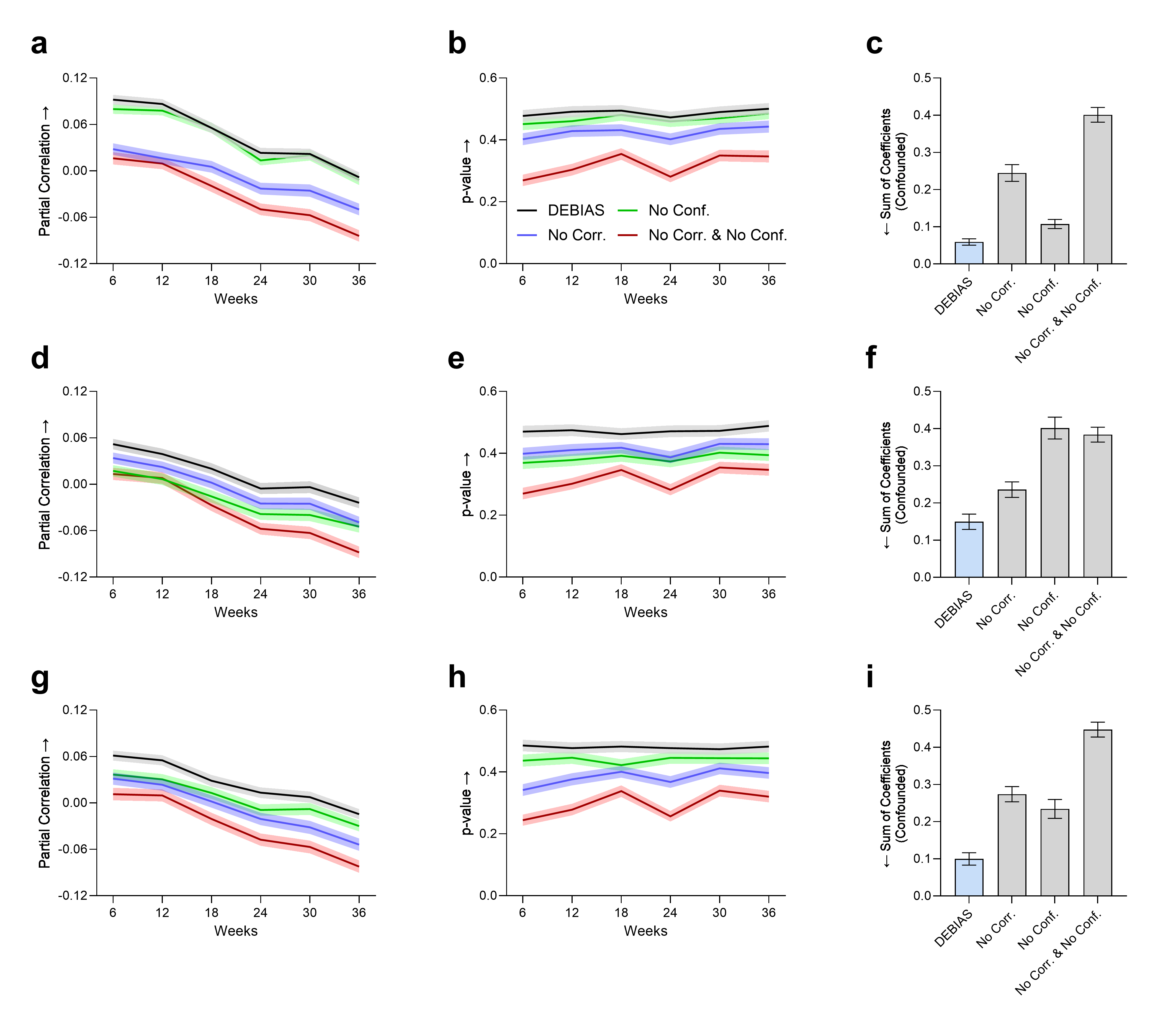}
    \caption{\textbf{Ablation accuracy results for TADS.} We evaluated DEBIAS against three ablated variants: replacing the correlation objective with MSE (no corr.), removing the confounding penalty by setting $\lambda = 0$ (no conf.), and both replacing the correlation with MSE and removing the confounding penalty (no corr. and no conf.). The first two columns of subfigures share the same legend in panel (b). For the first recovered score, panels (a) - (c) show that DEBIAS achieved the highest correlation with the learned outcomes (a), the highest p-value for the confounding tests (b), and the lowest total weight assigned to confounded outcome items (c). DEBIAS similarly outperformed all ablated variants for the second (d-f) and third (g-i) recovered scores across all metrics. As expected, the variant with both modifications (no corr. and no conf.) consistently performed the worst across all measures. These results indicate that both correlation maximization and confounding minimization are essential for DEBIAS to achieve optimal performance.}
    \label{fig:TADS:ablation}
\end{figure}

\newpage
\subsection{Additional CATIE Results}

\end{document}